%% file: ijcai24.tex
\documentclass{article}
\usepackage{ijcai24}

\usepackage{times}
\usepackage{soul}
\usepackage{url}
\usepackage[hidelinks]{hyperref}
\usepackage[utf8]{inputenc}
\usepackage[small]{caption}
\usepackage{graphicx}
\usepackage{amsmath}
\usepackage{amsthm}
\usepackage{booktabs}
\usepackage{algorithm}
\usepackage{algorithmic}
\usepackage[switch]{lineno}
\usepackage{amssymb}
\usepackage{array}
\usepackage{color}
\usepackage{multirow}
\usepackage{natbib}

\usepackage{color}
\usepackage{float}

\definecolor{green(pigment)}{rgb}{0.1607, 0.3843, 0.0941}
\definecolor{blue(pigment)}{rgb}{0., 0.1484, 0.6992}
\hypersetup{hidelinks}
\hypersetup{colorlinks,linkcolor=blue(pigment),citecolor=green(pigment),urlcolor=black}

\title{Preference Alignment on Diffusion Model: A Comprehensive \\ Survey for Image Generation and Editing}

\author{
Sihao Wu$^1$
\and
Xiaonan Si$^2$\and
Chi Xing$^{3}$\and
Jianhong Wang$^4$\and
Gaojie Jin$^5$\and\\
Guangliang Cheng$^1$\and
Lijun Zhang$^2$\and
Xiaowei Huang$^1$\and
\\
\affiliations
$^1$University of Liverpool, $^2$Institute of Software Chinese Academy of Sciences\\
$^3$University of Edinburgh, 
$^4$University of Bristol, $^5$University of Exeter\\
\emails
\{sihao.wu, guangliang.cheng, xiaowei.huang\}@liverpool.ac.uk,
\{sixiaonan, zhanglj\}@ios.ac.cn,
s2682783@ed.ac.uk, jianhong.wang@bristol.ac.uk, g.jin@exeter.ac.uk
}

\begin{document}

\maketitle

\begin{abstract}

The integration of preference alignment with diffusion models (DMs) has emerged as a transformative approach to enhance image generation and editing capabilities. Although integrating diffusion models with preference alignment strategies poses significant challenges for novices at this intersection, comprehensive and systematic reviews of this subject are still notably lacking. To bridge this gap, this paper extensively surveys preference alignment with diffusion models in image generation and editing. First, we systematically review cutting-edge optimization techniques such as reinforcement learning with human feedback (RLHF), direct preference optimization (DPO), and others, highlighting their pivotal role in aligning preferences with DMs. Then, we thoroughly explore the applications of aligning preferences with DMs in autonomous driving, medical imaging, robotics, and more. Finally, we comprehensively discuss the challenges of preference alignment with DMs. To our knowledge, this is the first survey centered on preference alignment with DMs, providing insights to drive future innovation in this dynamic area.

\end{abstract}

\input{contents/01_introduction}

\input{contents/02_preliminaries}

\input{contents/04_dm_rl_training}

\input{contents/05_dm_rl_applications}

\input{contents/09_future_directions}

\input{contents/10_conclusion}





\bibliographystyle{named}
\bibliography{ijcai24}

\twocolumn
\newpage




\end{document}

%% file: contents/01_introduction.tex
\section{Introduction}
\par \noindent Advancements in Generative AI have greatly enhanced image generation and editing with diffusion models (DMs) ~\citep{ramesh2022hierarchical,saharia2022photorealistic}, addressing key challenges across applications such as media enhancement~\citep{zhou2024surveygenerativeaillm}, driving simulation~\citep{guan2024world}, and virtual reality~\citep{10765093}. This progress broadens visual media capabilities and optimizes the production and integration of digital content on diverse platforms. Despite these advances, images generated or edited with DMs often encounter issues such as text-image misalignment, deviations from human aesthetic preferences, and propagation of content that may include toxic or biased content. 

To address these challenges, scholars have employed reinforcement learning (RL) strategies, notably using reward models (RM) to infer human preferences from expert-annotated output. A key instance is the Large Language Model (LLM), which applies reinforcement learning with human feedback (RLHF) to steer the intention towards human values and preferences \citep{achiam2023gpt}. In image generation and editing, many initiatives adopt various preference alignment strategies to produce images that better align with textual descriptions, aesthetics, and human values. However, the various forms of integration between DMs and preference alignment strategies pose challenges for novices in understanding the intersection of these domains. Therefore, this paper addresses the gap by systematically reviewing how DMs and RL integrate for image generation and editing, offering a concise overview and anticipating future developments. 

Although there are numerous surveys on diffusion-based image generation~\citep{DBLP:journals/corr/abs-2303-07909} and image editing~\citep{DBLP:journals/corr/abs-2402-17525}, few have provided a systematic and comprehensive review that thoroughly integrates DMs with RL in the field of image generation and editing. Several studies~\citep{DBLP:journals/corr/abs-2407-13734,du2023beyond,DBLP:journals/corr/abs-2308-14328} have reviewed the integration of text-to-image models with RL; however, they tend to primarily concentrate on the applications of RL to diffusion processes or offer overarching assessments of generative AI applications. They do not systematically explore the specific integration of DMs and RL in image generation and editing, nor adequately summarize the prevailing challenges and prospective direction in this field. To address this gap, this paper provides a comprehensive review of preference alignment-based DMs, examining their theoretical foundations, developmental progress, and practical implementations in the realms of image generation and editing. Building on these insights, it identifies key challenges and outlines future research directions, offering a forward-looking perspective on potential advancements in this rapidly evolving field.

Our contributions include: i) systematically reviewing optimization techniques such as RLHF, DPO, and others, highlighting their role in the preference alignment with DMs (Sec.~\ref{sec:preference_fine_tuning_on_t2i}); ii) thoroughly exploring applications of DMs in autonomous driving, medical imaging, robotics, and more (Sec.~\ref{sec:applications}); iii) comprehensively discussing challenges of preference alignment with DMs (Sec.~\ref{sec:challenges_and_future_directions}). To our knowledge, this paper is the first survey specifically focusing on preference alignment with DMs in image generation and editing, aiming to help researchers deepen their understanding and drive further innovation.

%% file: contents/02_preliminaries.tex
\section{Preliminaries}


\subsection{Diffusion Model}
\label{sec:pre_diffusion_model}

DDPMs employ a forward diffusion process that iteratively adds Gaussian noise to the data \( x_0 \) over \( T \) time steps. This process forms a Markov chain, where the data at each time step \( x_t \) depend only on the data from the immediately preceding step \( x_{t-1} \)~\citep{ho2020denoising}. The forward process is governed by the following equation:
\begin{equation}
\begin{aligned}
q(x_{t}\mid x_{t-1})=\mathcal N(x_{t};\sqrt{1-\beta_{t}}x_{t-1},\beta_{t} I),\nonumber
\end{aligned}
\end{equation}
with noise schedules ($\beta_{1}, ..., \beta_{T} $)  determining how quickly the original data degrades. The distribution at any step  $t$ can also be computed directly as following form.
\begin{equation}
\begin{aligned}
q(x_{t}\mid x_{0}) = \mathcal N (x_{t};\sqrt{ \bar{\alpha_{t}}}x_{0},(1- \bar{\alpha_{t}})I),\nonumber
\end{aligned}
\end{equation}
During the denoising phase, the reverse process  starts from pure noise at time step  $T$, uses a learnable model to iteratively denoise and reconstruct the original signal $x_{0}$, which is formulated as below:
\begin{equation}
\begin{aligned}
p_{\theta } (x_{t-1} \mid x_{t}) = \mathcal N (x_{t-1}; \mu_{\theta }  (x_{t}, t),  {\textstyle \sum_{\theta}} (x_{t},t)),\nonumber
\end{aligned}
\end{equation}
The model is trained by minimizing a variational lower bound on the negative log-likelihood of the data.
{\small
\begin{equation}
\begin{aligned}
\mathbb {E}[-\log p_{\theta }(x_{0})] \le  \mathbb{E}_{q}\left[ -\log p(x_{T})-\sum_{ t>1}\log \frac{p_{\theta}(x_{t-1}\mid x_{t})}{q(x_{t}\mid x_{t-1})}   \right]. \nonumber
\end{aligned}
\end{equation}
}

DDIMs is a more efficient and deterministic alternative, replacing the Markov chain with a non-Markovian process~\citep{song2020denoising}. This approach enables faster generation, accurate reconstruction, and supports applications such as image and video editing. The backward denoising step can be computed as follows:
{\small
\begin{equation}
\label{eq:back_denoise}
\begin{aligned}
x_{t-1} =\sqrt{ \bar {\alpha}_{t-1}}\frac{x_{t}-\sqrt{1-\alpha_{t} }\epsilon _{\theta }(x_{t},t)  }{\sqrt{\alpha_{t}}} + \sqrt{1-\alpha_{t-1} }\epsilon _{\theta }(x_{t},t). \nonumber
\end{aligned}
\end{equation}
}

\subsection{Reinforcement Learning}

The preference fine-tuning pipeline typically begins with a supervised fine-tuning (SFT) stage, where the diffusion model is trained to accurately map textual prompts to images. During SFT, the focus is on improving the model’s ability to produce high-quality images that closely reflect the prompt. Once the model is sufficiently proficient at generating images, it is further refined via reinforcement learning (RL). Proximal Policy Optimization (PPO) \citep{schulman2017proximal} is commonly employed for this step, owing to its stability and robustness during training. The following section provides a detailed overview of PPO in this context.


\noindent \textbf{PPO.} In each update timestep, the PPO algorithm collects a batch of transition samples using a rollout policy $\pi_{\theta_{\text{old}}}(a_t | s_t)$ and optimizes a clipped surrogate objective, as follows:
{\small
\begin{align}
L^{\text{CLIP}}(\theta) = \hat{\mathbb{E}}_t \left[ \min \left( r_t(\theta) A_t, \, \text{clip}\left( r_t(\theta), 1 - \epsilon, 1 + \epsilon \right) A_t \right) \right], \nonumber
\end{align}
}

\noindent where $r_t(\theta)$ represents the probability ratio, defined as $r_t(\theta) = \frac{\pi_\theta(a_t \mid s_t)}{\pi_{\theta_{\text{old}}}(a_t \mid s_t)}$. 
The term $A_t$ is the advantage in the timestep $t$, estimated using Generalized Advantage Estimation (GAE)~\citep{schulman2015high}, 
and $\epsilon$ is a hyperparameter that controls the width of the clipping interval. The value function is trained by minimizing the following loss function:
{\small
\begin{align}
L^V(\theta) = \hat{\mathbb{E}}_t \big[ &\frac{1}{2} \max \big( 
\left( V_\theta - V_t^{\text{targ}} \right)^2, \nonumber
\\
&\left( \text{clip}(V_\theta, V_{\text{old}} - \epsilon, V_{\text{old}} + \epsilon) - V_t^{\text{targ}} \right)^2 
\big) \big], \nonumber
\end{align}
}

\noindent where $V_t^{\text{targ}}$ denotes the bootstrapped value function target. Additionally, an entropy bonus is often incorporated to encourage sufficient exploration such that
\begin{equation}
\begin{aligned}
L^H(\theta) &= \hat{\mathbb{E}}_t \left[ H[\pi_\theta](s_t) \right], \nonumber
\end{aligned}
\end{equation}

\noindent where $H[\pi_\theta](s_t)$ is the entropy of a policy $\pi_\theta$ at state $s_t$. Incorporating the above loss functions, the overall optimization objective of PPO is defined as:
\begin{equation}
\begin{aligned}
L^{\text{PPO}}(\theta) &= -L^{\text{CLIP}}(\theta) + \lambda_V L^V(\theta) - \lambda_H L^H(\theta), \nonumber
\end{aligned}
\end{equation}

\noindent where $\lambda_V$ and $\lambda_H$ are weighting coefficients to balance the importance of the value function loss and the entropy bonus. The algorithm alternates between collecting trajectory data using the policy and optimizing the collected data based on this loss function, until convergence.

%% file: contents/04_dm_rl_training.tex
\section{Preference Alignment on DMs}
\label{sec:preference_fine_tuning_on_t2i}
\subsection{Framework}

\begin{figure*}[th!]
    \centering
    \includegraphics[width=1\linewidth]{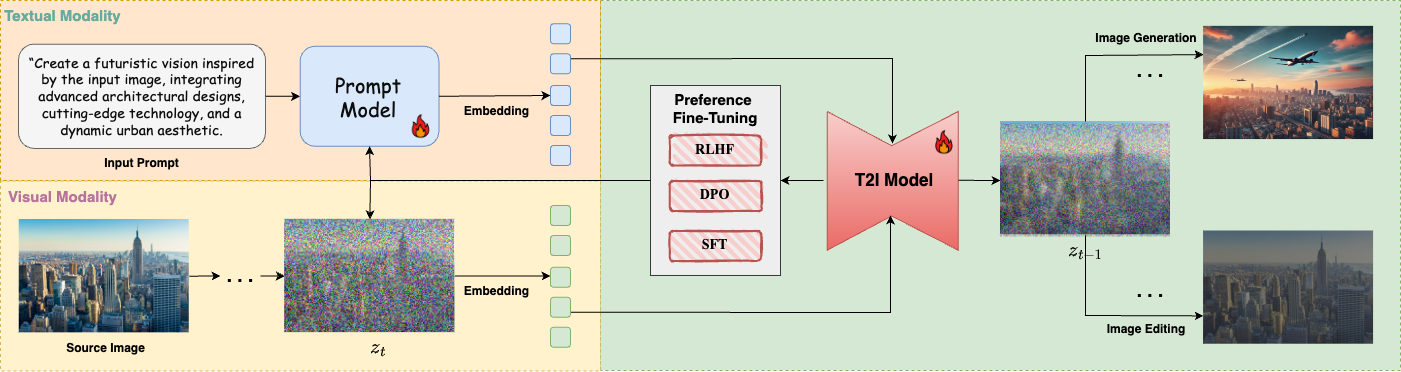}
    \caption{Preference Alignment on DM Framework: The process begins with a prompt model encoding a prompt into an embedding, and a source image is transformed into a latent representation $z_t$. DMs fine-tuned with RLHF, DPO, and SFT, process these embeddings to generate new images or edit existing ones while preserving structural integrity, ensuring outputs align with user preferences.}
    \label{fig:total_framework}
\end{figure*}

\begin{table*}[th!]
\centering
\resizebox{1\linewidth}{!}
{\input{tables/03_image_generation_editing_classification}}
\caption{Preference Alignment methods, which are categorized based on the approaches under study, without restricting their applicability to other domains or modalities.}
\label{tab:method_categories}
\end{table*}

Generally, the preference alignment process for generative models can be formulated as an RL problem, which includes the following components: a policy model, a reward model, an action space, and an environment. The policy model $\pi_\theta$ is a generative model taking an input as prompt $x$, and producing either a sequence of outputs or probability distributions $y$. Furthermore, we categorize all preference alignment approaches into the following taxonomy, as shown in Table~\ref{tab:method_categories}.

\noindent \textbf{Sampling.} In RL, methods are categorized by how data is sampled and used to train or obtain rewards: online and offline human alignment. Online alignment refers to generative models acting as policies that collect data through interaction with their environments, acquiring reward values either from a predefined reward model or from the policies themselves. Online methods are further divided into on-policy learning with identical behavior and target policies and off-policy learning with different behavior and target policies. Offline alignment means that generative models use collected human demonstrations in prior for training.

\noindent \textbf{Training Strategy.} Training strategies for preference alignment in DMs include RL, DPO, and others. RL techniques, such as PPO, allow models to align with human preferences by maximizing a reward function. DPO directly optimizes the reward signal without relying on RL steps, providing a computationally efficient alternative that will be detailed later. We discuss GFlowNets-based and self-play methods in others.

\noindent \textbf{Reward Feedback.} Data collection strategies can be divided into two main genres: human annotation and AI-generated annotation. Human annotation means collecting preference data through manual evaluations and comparisons, which provide high-quality but often limited datasets. AI annotation leverages pre-trained models to generate large-scale annotations automatically, serving as a scalable and cost-effective approach. These two methods are often combined to create hybrid datasets that balance quality and scalability.

\noindent \textbf{Modalities.} The approaches reviewed are categorized based on their optimized modalities: textual, visual, and multi-modalities. Textual modalities focus on optimizing textual prompts for improved image generation. Visual modalities mean directly updating the DM. Multi-modalities approaches combine textual and visual modalities to create comprehensive frameworks that leverage cross-modal information, resulting in more cohesive and context-aware outputs.

\subsection{RLHF}
Generally, RLHF involves learning a reward function from human feedback and subsequently optimizing policies with respect to the reward~\citep{christiano2017deep}. The training process for RLHF consists of three stages. Firstly, the policy model \( \pi_\theta \) interacts with the environment, and its parameters are updated using reinforcement learning techniques. Next, pairs of output segments generated by \( \pi_\theta \) are selected and presented to human annotators for preference comparison. Finally, the model's parameters are optimized based on a reward signal \( r \), which is elicited from the human-provided comparisons.

As briefly introduced in Sec.~\ref{sec:pre_diffusion_model}, the formulation of text-to-image DMs can be described as a Markov Chain Process (MDP). There are two typical related works in RLHF for DMs, DDPO \citep{black2023training} and DPOK \citep{fan2024reinforcement}. Consider taking $(i, x_i, c)$ as the state space, and define the action as the next hierarchy $x_{i-1}$ to go to. Then, Eq.~\ref{eq:back_denoise} naturally defines a stochastic policy: the stochasticity of the policy comes from $\sqrt{\beta_i z_i}$. Accordingly, the policy follows a Gaussian distribution $\mathcal{N} \big(s_\theta^* (i, x_i, c), \beta_i \big)$ with the mean determined by $s_\theta^* (i, x_i, c)$ and variance $\beta_i$:
\begin{equation}
\begin{aligned}
\pi_\theta(x_{i-1} \mid x_i) \sim  \mathcal{N} \left(\frac{1}{\sqrt{1 - \beta_i}} \left( x_i + \beta_i s_\theta(i, x_i, c) \right), \beta_i \right). \nonumber
\end{aligned}
\end{equation}

Given this formulation, DDPO \citep{black2023training} directly maximize the expected reward, 
$J_{\text{DDPO}} = \mathbb{E}_\theta [r(x_0, c)],$
using either the REINFORCE or PPO algorithms, without applying regularization.
\vspace{-5pt}
\begin{equation}
\begin{aligned}
\nabla_\theta J_{\text{DDPO}} = \mathbb{E} \left[ 
\sum_{t=0}^{T} \nabla_\theta \log p_\theta(x_{t-1} \mid x_t, c) \, r(x_0, c)
\right]. \nonumber
\end{aligned}
\end{equation}

In contrast, DPOK \citep{fan2024reinforcement}, optimizes the regularized reward objective as below. They proposed a clipped gradient algorithm inspired by the PPO objective and showed that incorporating regularization improves generation quality over unregularized versions.
\begin{equation}
\resizebox{.95\linewidth}{!}{$
\begin{aligned}
J_{\text{DPOK}} = \mathbb{E}_\theta \big[r(x_0, c)\big] 
- \beta \mathbb{E}_{p(z)} \big[\text{KL}(p_\theta(x_0 \mid z) \parallel p_{\text{pre}}(x_0 \mid z))\big]. \nonumber
\end{aligned}$}
\end{equation}

Moreover, the Human Preference Score (HPS) aligns T2I models with human preferences by leveraging a classifier trained on a large-scale dataset of human choices~\citep{wu2023human}. \citet{xu2024imagereward} applies Reward Feedback Learning (ReFL) for better synthesis, which trains on 137,000 expert comparisons. VersaT2I~\citep{guo2024versat2i} adopts a self-training approach with Low-Rank Adaptation (LoRA) to address multiple image quality aspects, introducing a mixture of LoRA models for improved quality consistency. The Parrot framework \citep{lee2025parrot} employs a multi-reward optimization algorithm using reinforcement learning to balance aesthetics, alignment, sentiment, and human preferences through Pareto-optimal selection, achieving superior image quality. Diffusion-MVP integrates visual rewards into text-to-image generation including Visual Market, Visual Aesthetic, and CLIP relevance~\citep{he2023learning}.  To address the depth-efficiency trade-off, Deep Reward Tuning (DRTune) optimizes reward functions in DMs by controlling early denoising steps~\citep{wu2025deep}. LRLD~\citep{zhang2025large} introduces a scalable RL framework that focuses on improving text-to-image alignment with human preferences, fairness, and object composition across a
diverse set of reward functions. 

On the other hand, recent efforts in fine-tuning DMs have expanded to \textbf{image editing}. InstructRL4Pix \citep{li2024instructrl4pix} introduces a reinforcement learning framework for image editing, utilizing attention maps and Proximal Policy Optimization to enhance feature localization and preservation. Incorporating human preferences into image editing workflows, HIVE takes advantage of human feedback to train a reward function~\citep{zhang2024hive}. To address the challenge of generative diversity, DRLDiv \citep{miao2024training} is a method that quantifies the overlap between generative and reference distributions. LfVoid exemplifies multi-modal integration by leveraging pre-trained T2I models to generate visual objectives for robotic learning tasks~\citep{gao2023can}. By editing images based on natural language instructions, it guides RL agents without in-domain data, showcasing strong performance in simulated and real-world tasks and the potential of generative models in robotics.

The above methods underscore the growing importance of human-centered optimization strategies in T2I generation, while further advancements have introduced exploration into \textbf{AI feedback} mechanisms. Specifically, LLMs and LVLMs have demonstrated to function with minimal human intervention as the evaluation strategy. 
\citet{liang2024rich} introduces RAHF, which provides automatic feedback on implausible regions and misaligned keywords, facilitating training data refinement and region inpainting. LVLM-REFL integrates LVLMs to evaluate alignment between generated images and input texts~\citep{wen2023improving}. Additionally, this method incorporates an iterative correction process during inference, applying image-editing algorithms guided by the LVLM.    

Recently, there has been a growing surge of interest in the precise refinement of \textbf{text modality}. Prompt Auto-Editing (PAE) utilizes a two-stage training process combining supervised fine-tuning and online RL for dynamic prompts control~\citep{mo2024dynamic}. This approach enables the iterative refinement of prompts of the weights and injection time steps of each word. In contrast, NegOpt optimizes negative prompts to improve image aesthetics and fidelity~\citep{ogezi2024optimizing}. 
TextCraftor~\citep{li2024textcraftor} focuses on the text encoder, leveraging reward functions to optimize performance without paired datasets. 
The integration of LLMs and VLMs as carriers or alternative representations of reward functions has introduced novel strategies for optimizing textual modality. OPT2I~\citep{manas2024improving} exemplifies this by employing automatic prompt optimization, where an LLM iteratively refines prompts to maximize a consistency score while preserving image quality and improving recall between generated and real data. Similarly, the RPG framework~\citep{yang2024mastering} advances text-to-image generation by using multimodal LLMs to decompose complex prompts into detailed subprompts, allocate these across image regions via chain-of-thought planning, and apply complementary regional diffusion for precise image synthesis.
DiffChat introduces an interactive image creation framework that leverages the InstructPE dataset and RL to integrate feedback on aesthetics, user preferences, and content integrity, enabling efficient generation of high-quality target prompts~\citep{wang2024diffchat}. These approaches underscore the growing importance of LLMs and VLMs in enhancing prompt optimization and aligning text-to-image models with user-driven goals.

Another common approach is direct reward backpropagation. Direct reward backpropagation simplifies implementation, especially with a pre-trained differentiable reward model, and accelerates training by directly backpropagating reward gradients. DRaFT propagates gradients directly from reward functions to update policies~\citep{clark2023directly}. It has proven effective across a variety of reward functions, significantly enhancing the aesthetic quality of images generated by Stable Diffusion.  \citet{prabhudesai2024aligningtexttoimagediffusionmodels} introduces AlignProp, reformulating denoising in text-to-image DMs as a differentiable recurrent policy. It uses randomized truncated backpropagation to mitigate over-optimisation, randomly selecting the denoising step for reward backpropagation.
Furthermore, \citet{uehara2024finetuningcontinuoustimediffusionmodels} propose ELEGANT to directly optimize entropy-enhanced rewards with neural SDEs, effectively preventing excessive reward maximization.

\subsection{DPO and its Variants}
This section explains each preference alignment method like DPO and its variants. One notable drawback of RLHF is that the RL optimization step often demands significant computational resources \citep{winata2024preference}, such as those required for PPO. To address this, DPO, recently proposed by~\citep{rafailov2024direct}, offers a promising alternative by bypassing the reward modelling stage and eliminating the need for reinforcement learning, which has garnered considerable attention. The core idea of DPO is based on the observation that, given a reward function $r(x, y)$, 
the DPO objective can be described as follows:
\begin{equation}
\resizebox{.95\linewidth}{!}{$
\begin{aligned}
\mathcal{L}_{\text{DPO}}(\pi_\theta; \pi_{\text{ref}}) := & - \mathbb{E}_{(x, y_w, y_l) \sim \mathcal{D}} \nonumber \\
& \Bigg[\log \sigma \Bigg( 
\beta_{\text{reg}} \log \frac{\pi_\theta(y_w \mid x)}{\pi_{\text{ref}}(y_w \mid x)} - \beta_{\text{reg}} \log \frac{\pi_\theta(y_l \mid x)}{\pi_{\text{ref}}(y_l \mid x)}
\Bigg)
\Bigg]. \nonumber
\end{aligned}$}
\end{equation}

\citet{wallace2024diffusion} investigates human preference learning in text-to-image DMs and introduces Diffusion-DPO, a novel approach to aligning these models with user preferences. By adapting the DPO framework, Diffusion-DPO reformulates the objective of incorporating the evidence lower bound, enabling a differentiable and efficient optimization process. HF-T2I ~\citep{lee2023aligning} proposes a reward-weighted likelihood maximization framework by collecting binary feedback. As for temporal rewards DPO, \citet{yang2024dense} introduces a dense reward perspective with temporal discounting in DPO-style objectives, prioritizing early steps in the generation process to improve alignment efficiency. PRDP~\citep{deng2024prdp} reformulates RLHF as a supervised regression task, allowing stable fine-tuning in large prompt datasets. In response to the provided text prompts, the method selects two image candidates and trains the DM to predict reward differences between them based on their denoising trajectories. The model parameters are updated based on the prediction's mean square error loss. In DMs, the storage of gradients for multiple latent image representations, which are substantially larger than word embeddings, imposes memory requirements that are frequently infeasible. In \citep{yang2024using} (D3PO), the denoising process is reframed as a multi-step MDP to reduce computational overhead and enable DPO-based fine-tuning of DMs with human feedback. Using a pre-trained model to represent the action value function  $Q$, the DPO framework is extended to allow direct parameter updates at each denoising step, eliminating the need for a reward model and significantly reducing computational costs. Diffusion-KTO introduces a novel approach for aligning text-to-image models by using only per-image binary feedback, removing the need for costly pairwise preference data \citep{li2024aligningdiffusionmodelsoptimizing}. TailorPO~\citep{ren2025refiningalignmentframeworkdiffusion} ranks intermediate noisy samples by step-wise reward and efficiently addresses gradient direction issues.

Unlike DPO, which operates on a fixed offline dataset, \textbf{Online DPO} dynamically constructs preference datasets by leveraging model outputs and various labelers, such as pre-trained reward models, LLM judges, or even the trained model itself via prompting. To address off-policy optimization and scalability challenges in existing video preference learning frameworks, \citet{zhang2024onlinevpoalignvideodiffusion} propose an online DPO algorithm (OnlineVPO), which continually refines video generation quality by incorporating feedback in real time. Moreover, Iterative Preference Optimization (IPO) extends the concept of integrating human feedback to enhance generated video quality \citep{yang2025ipoiterativepreferenceoptimization}. Specifically, IPO utilizes a critic model to evaluate and rank video outputs pairwise, similar to the approach used in DPO, thereby refining the generative process based on explicit preference signals.  

\subsection{Others}

In addition to RL and DPO, some literature uses GFlowNets~\citep{bengio2023gflownet}, also coined as the consistency model, to achieve direct reward backpropagation and alignment. Regarding the GFlowNet-based algorithm, the formulation of DM is directly connected to the definition of GFlowNet MDP, as highlighted in \citep{zhang2022unifying}. Importantly, the conditional distribution of the denoising process, $p_\theta(\mathbf{x}_{T-t-1} \, | \, \mathbf{x}_{T-t})$, aligns with the GFlowNet forward policy $P_F(s_{t+1} \, | \, s_t)$. Similarly, the conditional distribution of the diffusion process, $q(\mathbf{x}_{T-t} \, | \, \mathbf{x}_{T-t-1})$, corresponds to the GFlowNet backward policy $P_B(s_t \, | \, s_{t+1})$ \citep{lahlou2023theory}. Building on the above works, 
\citet{zhang2024improving} proposes the
DAG to post-train DMs with
black-box property functions. Further, they propose a KL-based objective for optimizing GFlowNets to offer improved sample efficiency. Moreover, RLCM proposes a framework for fine-tuning consistency models via RL, framing the iterative inference process as an MDP~\citep{oertell2024rl}. This approach speeds up training and inference while maintaining the efficiency of consistency models, enhancing performance across metrics. However, as a caveat, this property might also pose challenges in effectively generating high-reward samples beyond the training data. This implies that these approaches may not be suitable when accurate reward feedback is readily available without learning. Hence, we generally recommend using them when reward functions are unknown. Innovative self-play methods like SPIN-Diffusion~\citep{yuan2024self} remove the need for human preference data, achieving superior performance in text-to-image tasks. By engaging the model in a self-improvement process, where the diffusion
model engages in competition with its earlier versions, facilitating an iterative self-improvement
process. SPIN-Diffusion ensures efficient fine-tuning that leads to improved alignment with target distributions.

%% file: tables/03_image_generation_editing_classification.tex
\begin{tabular}{@{}lccccccccc@{}}
\toprule
\textbf{Method}   & \multicolumn{3}{c}{\textbf{Training Strategy}}  & \multicolumn{2}{c}{\textbf{Task}} & \multicolumn{2}{c}{\textbf{Reward Feedback}} & \multicolumn{2}{c}{\textbf{Modality}} 
\\
\cmidrule(lr){2-4} \cmidrule(lr){5-6} \cmidrule(lr){7-8} \cmidrule(lr){9-10}
& DPO & RLHF & Others  & Generation & Editing  & Human & AI Feedback & Textual & Visual \\
\midrule
Online Methods \\ \midrule 

DDPO \citep{black2023training} &  & \checkmark &   & \checkmark &  & \checkmark &  &  & \checkmark \\
DPOK \citep{fan2024reinforcement} &  & \checkmark &   & \checkmark &  & \checkmark &  &  & \checkmark \\
HPS \citep{wu2023human} &  & \checkmark &   & \checkmark &  & \checkmark &  &  & \checkmark \\
ReFL  \citep{xu2024imagereward} &  & \checkmark &   & \checkmark &  & \checkmark &  &  & \checkmark \\
VersaT2I  \citep{guo2024versat2i} &  & \checkmark &   & \checkmark &  & \checkmark &  &  & \checkmark \\
Parrot \citep{lee2025parrot} &  & \checkmark &   & \checkmark &  & \checkmark &  &  & \checkmark \\
Diffusion-MVP  \citep{he2023learning} &  & \checkmark &   & \checkmark &  & \checkmark &  &  & \checkmark \\
DRTun \citep{wu2025deep} &  & \checkmark &   & \checkmark &  & \checkmark &  &  & \checkmark \\
LRLD \citep{zhang2025large} &  & \checkmark &   & \checkmark &  & \checkmark &  &  & \checkmark \\

InstructRL4Pix \citep{li2024instructrl4pix}&  & \checkmark &  &  & \checkmark & \checkmark &  &  & \checkmark \\
HIVE \citep{zhang2024hive}&  & \checkmark &  &  & \checkmark & \checkmark &  &  & \checkmark \\
DRLDiv \citep{miao2024training}&  & \checkmark &  &  & \checkmark & \checkmark &  &  & \checkmark \\
LfVoid \citep{gao2023can} &  & \checkmark &  &  & \checkmark & \checkmark &  & \checkmark & \checkmark \\

RAHF \citep{liang2024rich} &  & \checkmark &   & \checkmark &  &  & \checkmark  &  & \checkmark \\
LVLM-REFL \citep{wen2023improving} &  & \checkmark &   & \checkmark &  &  & \checkmark &  & \checkmark \\

PAE \citep{mo2024dynamic} &  & \checkmark &   & \checkmark &  &  &  & \checkmark & \\
NegOpt  \citep{ogezi2024optimizing} &  & \checkmark &   & \checkmark &  & \checkmark &  & \checkmark & \\
TextCraftor  \citep{li2024textcraftor} &  & \checkmark &   & \checkmark &  & \checkmark &   & \checkmark & \\
OPT2I \citep{manas2024improving} &  & \checkmark &   & \checkmark &  &  & \checkmark & \checkmark & \\
RPG \citep{yang2024mastering} &  & \checkmark  &   & \checkmark &  &  & \checkmark & \checkmark & \\
DiffChat \citep{wang2024diffchat} &  & \checkmark &   & \checkmark &  &  & \checkmark  & \checkmark & \\

DRaFT \citep{clark2023directly} &  & \checkmark &  & \checkmark &  & \checkmark &  &  & \checkmark \\
AlignProp \citep{prabhudesai2024aligningtexttoimagediffusionmodels} &  & \checkmark &  & \checkmark &  & \checkmark &  &  & \checkmark \\
ELEGANT \citep{uehara2024finetuningcontinuoustimediffusionmodels} &  & \checkmark &  & \checkmark &  & \checkmark &  &  & \checkmark \\

IPO \citep{yang2025ipoiterativepreferenceoptimization} & \checkmark & & & \checkmark & & \checkmark & & & \checkmark \\
OnlineVPO \citep{zhang2024onlinevpoalignvideodiffusion} & \checkmark & & & \checkmark & & \checkmark & & & \checkmark \\

DAG \citep{zhang2024improving} &  & & \checkmark & \checkmark & & \checkmark & & & \checkmark \\
RLCM \citep{oertell2024rl} &  & & \checkmark & \checkmark & & \checkmark & & & \checkmark \\
SPIN-Diffusion \citep{yuan2024self} &  & & \checkmark & \checkmark & & \checkmark & & & \checkmark \\

\midrule
Offline Methods \\ \midrule

Diffusion-DPO \citep{wallace2024diffusion} & \checkmark &  &  & \checkmark &  & \checkmark &  &  & \checkmark \\
HF-T2I \citep{lee2023aligning} & \checkmark &  &  & \checkmark &  & \checkmark &  & \checkmark & \checkmark \\
temporal-DPO \citep{yang2024dense} & \checkmark & & & \checkmark & & \checkmark &  &  & \checkmark \\
PRDP \citep{deng2024prdp} & \checkmark &  & \checkmark & \checkmark &  & \checkmark &  & \checkmark & \checkmark \\
D3PO \citep{yang2024using} & \checkmark & & & \checkmark & &\checkmark & & & \checkmark \\
Diffusion-KTO \citep{li2024aligningdiffusionmodelsoptimizing} & \checkmark & & & \checkmark & &\checkmark & & & \checkmark \\
TailorPO \citep{ren2025refiningalignmentframeworkdiffusion} & \checkmark & & & \checkmark & &\checkmark & & & \checkmark  \\

\bottomrule
\end{tabular}

%% file: contents/05_dm_rl_applications.tex
\section{Applications}
\label{sec:applications}


\begin{figure*}[t]
    \centering
    \includegraphics[width=1\linewidth]{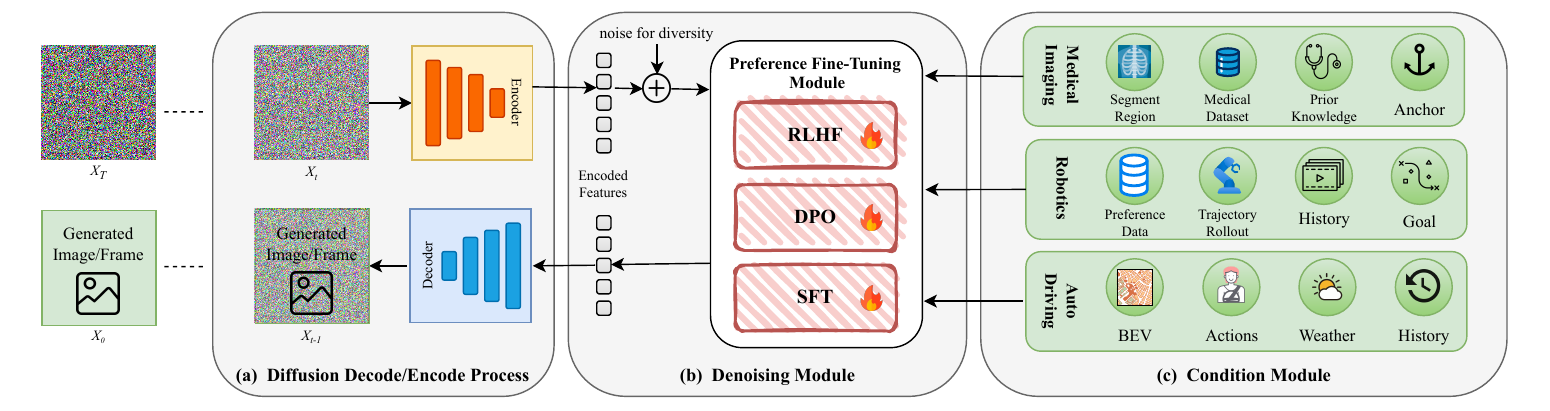}
    \caption{General paradigm of preference alignment for various \textbf{condition modules} on DM applications.}
    \label{fig:application_paradigm}
\end{figure*}
This section underscores the critical role of preference alignment in diffusion models across various domains, including medical imaging, robotics, and autonomous driving. As shown in Figure~\ref{fig:application_paradigm}, most applications rely on fine-tuning modules integrated into the denoising process, ensuring optimized generation conditioned on the specific requirements of each domain.
\subsection{Medical Imaging}

\textbf{Segmentation} Medical image segmentation partitions images into distinct regions of interest, playing a significant role in diagnosis and image-guided surgery. However, lesions and organs often exhibit ambiguity, making them difficult to be identified from the background. To this end, \citet{wu2022medsegdiff} proposed MedSegDiff to enhance the step-wise regional attention in DDPMs. Their approach optimizes the lesion’s feature map during the feature extraction phase in the diffusion encoder to enhance its visibility. Building on this work, \citet{wu2024medsegdiff} proposed MedSegDiff-V2, integrating diffusion processes and transformers with dual conditioning, anchor and semantic, guided by SS-Former to improve medical image segmentation. Furthermore, \citet{rahman2023ambiguous} utilized the stochastic sampling mechanism to generate a distribution of segmentation masks with cost-effective additional training.

\noindent\textbf{Synthesis \& Reconstruction} The scarcity of public medical imaging datasets, combined with strict privacy regulations, underscores the urgent need for advanced image synthesis techniques. \citet{daum2024differentially} presented latent DMs designed to generate synthetic images conditioned on  medical attributes while protecting patient privacy through differentially private training. These models are pre-trained on public datasets and then fine-tuned with differential privacy using the UK Biobank dataset. Subsequently, \citet{han2024advancing} fused RLHF strategies and proposed CXRL to optimize the generation result.
Moreover, preference fine-tuned DMs are widely used to reconstruct high-quality medical imaging data by integrating feature engineering into the denoising process~\citep{xie2022measurement}, effectively addressing issues such as missing data, noise interference, and low resolution.

\subsection{Robotics}

\textbf{Motion Generation} DMs have enhanced robustness and adaptability in motion generation by unifying offline learning with preference alignment and physics-informed control. Offline diffusion learning, combined with online preference alignment, improves quadrupedal locomotion by ensuring stability, precision, and zero-shot generalization to real-world robots~\citep{yuan2024preferencealigneddiffusionplanner}. These advancements showcase the power of DMs to capture complex motion dynamics in robotic motion learning. By generating diverse and realistic trajectories, these models enable robots to perform robust and adaptive behaviors, paving the way for more efficient and capable robotic systems.

\noindent\textbf{Trajectory Planning} DMs have significantly advanced trajectory planning in robotics by facilitating preference alignment and decision-making across complex, multi-objective scenarios. \citet{janner2022planningdiffusionflexiblebehavior} embedded the trajectory optimization process into the denoising process, enabling long-horizon planning while addressing alignment inconsistencies between conditions and generated trajectories. Preference-conditioned DMs further enhance multi-task decision-making by maximizing mutual information to improve trajectory alignment~\citep{yu2024regularized}. 

\noindent\textbf{Policy Learning \& Others} DMs are revolutionizing policy learning in robotics by tackling challenges like multi-task learning~\citep{wang2024sparse}, data scarcity~\citep{liu2024rdt}, and aligning policies with human preferences~\citep{dong2023aligndiff}. Furthermore, preference fine-tuned DMs demonstrate remarkable potential not only in advancing the design and control of soft robots~\citep{wang2023diffusebot}, but also in the training of sophisticated world simulators~\citep{yang2023learning}. These simulators leverage DMs to generate high-fidelity, diverse, and contextually accurate virtual environments, providing a robust framework for training and testing autonomous systems in a wide range of scenarios.

\subsection{Autonomous Driving}
\noindent\textbf{Data Generation} Autonomous driving models require a large amount of high-quality and complex training data. However, real data are costly to collect with restricted coverage, so synthesizing data through generative techniques becomes an effective alternative. To improve data quality, specifically spatio-temporal consistency and resolution, RLHF, SFT, and DPO have all been shown to be remarkably effective methods. Several methods have been proved to generate training data for multi-view videos~\citep{li2023drivingdiffusion}, panoramic videos~\citep{wen2024panacea}, etc. 

\noindent\textbf{Decision Making} Traditional auto-driving decision-making processes rely on predicting the future behavior of traffic participants, such as vehicles and pedestrians, and using deterministic or rule-based methods for path planning. However, these approaches are limited by predicting only a few possible scenarios, failing to account for the uncertainty in real-world traffics. To mitigate these issues, several methods are proposed to optimize the decision-making process by RLHF~\citep{huang2024gen}; to accelerate the decision-making process by shortening denoising process~\citep{liao2024diffusiondrive}; and to increase the security of the decision-making process by merging RLHF and SFT~\citep{liu2024ddm}.


\noindent\textbf{World Model} Similar to robotics, world models hold great promise in autonomous driving, where accurately predicting diverse movements is essential for making safe and effective driving decisions~\citep{feng2025survey}. Recently, generalized diffusion-based world models have attracted considerable attention~\citep{chen2024drivinggpt}.

\subsection{Others}

The above discussion demonstrates that preference alignment on DMs exhibits remarkable potential in domains with a critical demand for large-scale training data. Beyond the above domains, in the field of biology, DMs can not only predict protein structures~\citep{wu2024protein} but also optimize their stability and functionality, expediting experimental workflows. In game simulation~\citep{menapace2024promptable}, these models generate realistic virtual worlds, simulate complex agent behaviors, and streamline gameplay mechanics, ultimately enhancing user immersion and engagement. Furthermore, in the content creation sphere, DMs facilitate the production of high-quality, preference-aligned outputs—spanning digital art, music, and other creative mediums—satisfying diverse aesthetic and stylistic requirements~\citep{lin2023magic3d}. Taken together, these applications illustrate how preference alignment on DMs can adapt across multidisciplinary domains, opening up new avenues for innovation and collaboration.

%% file: contents/09_future_directions.tex
\section{Challenges and Future Directions}
\label{sec:challenges_and_future_directions}
The field currently faces several fundamental challenges. 
The predominant pressing is the computational burden of preference fine-tuning, particularly in RLHF approaches that require loads of resources for both training and inference. 
This is along with the challenge of collecting high-quality human preference data, which is both time-consuming and expensive. 
Besides, maintaining semantic consistency while optimizing for multiple diverse objectives---such as image quality, text alignment, and aesthetic appeal---remains difficult. 
Furthermore, the subjective nature of human preferences also makes it challenging to design effective reward functions that can reliably capture and reproduce these preferences in scale.

Looking ahead, several promising directions emerge for advancing the field. 
One key area is developing more efficient fine-tuning methods that can reduce computational requirements while maintaining performance. 
This includes exploring parameter-efficient approaches like LoRA and techniques that can minimize the need for extensive human feedback. 
Another crucial direction is improving the integration of multiple modalities, particularly in combining text and image understanding to achieve better alignment with human preferences. 
This could involve taking the use of LLMs to better interpret and implement human preferences in image generation. The field is also moving toward more practical applications. 
There is growing interest in adapting these techniques for specific domains like medical imaging, autonomous driving, and robotics. 
This application-oriented direction requires more robust evaluation metrics and reliable models for real-world scenarios. 
Also, there is an increasing emphasis on making fine-tuning more interpretable and controllable, for a better understanding of how preferences affect generation and enabling more fine-grained control over different aspects of the output, respectively.

As these models grow more powerful and widely deployed, safety and ethical considerations become increasingly important. This includes developing evaluation metrics and approaches, to ensure alignment with human values, preventing misuse, and maintaining privacy and security during the fine-tuning process. Success in these areas will be essential milestones toward the responsible advancement of preference fine-tuning in DMs.

%% file: contents/10_conclusion.tex
\section{Conclusion}

This survey provides a comprehensive review of preference alignment methods for DMs, focusing on integrating RL-related techniques like RLHF and DPO to enhance alignment with human preferences. We categorize approaches into sampling, training strategies, data collection, and modalities, as well as highlight their applications in medical imaging, autonomous driving and robotics. Despite the temporary success, challenges such as computational efficiency and preference modelling still remain. Future research should address these issues, and continue to explore scalable algorithms, multimodal capabilities, and ethical considerations to expand the full potential of generative AI.

%% file: ijcai24.bbl
\begin{thebibliography}{}

\bibitem[\protect\citeauthoryear{Achiam \bgroup \em et al.\egroup }{2023}]{achiam2023gpt}
Josh Achiam, Steven Adler, and et~al.
\newblock Gpt-4 technical report.
\newblock {\em ArXiv}, 2023.

\bibitem[\protect\citeauthoryear{Bengio \bgroup \em et al.\egroup }{2023}]{bengio2023gflownet}
Yoshua Bengio, Salem Lahlou, and et~al.
\newblock Gflownet foundations.
\newblock {\em JMLR}, 2023.

\bibitem[\protect\citeauthoryear{Black \bgroup \em et al.\egroup }{2023}]{black2023training}
Kevin Black, Michael Janner, and et~al.
\newblock Training diffusion models with reinforcement learning.
\newblock {\em ArXiv}, 2023.

\bibitem[\protect\citeauthoryear{Cao \bgroup \em et al.\egroup }{2023}]{DBLP:journals/corr/abs-2308-14328}
Yuanjiang Cao, Quan~Z. Sheng, and et~al.
\newblock Reinforcement learning for generative {AI:} {A} survey.
\newblock {\em Arxiv}, 2023.

\bibitem[\protect\citeauthoryear{Chen \bgroup \em et al.\egroup }{2024}]{chen2024drivinggpt}
Yuntao Chen, Yuqi Wang, and et~al.
\newblock Drivinggpt: Unifying driving world modeling and planning with multi-modal autoregressive transformers.
\newblock {\em ArXiv}, 2024.

\bibitem[\protect\citeauthoryear{Christiano \bgroup \em et al.\egroup }{2017}]{christiano2017deep}
Paul~F Christiano, Jan Leike, and et~al.
\newblock Deep reinforcement learning from human preferences.
\newblock {\em Neurips}, 2017.

\bibitem[\protect\citeauthoryear{Clark \bgroup \em et al.\egroup }{2023}]{clark2023directly}
Kevin Clark, Paul Vicol, and et~al.
\newblock Directly fine-tuning diffusion models on differentiable rewards.
\newblock {\em ArXiv}, 2023.

\bibitem[\protect\citeauthoryear{Daum and et al.}{2024}]{daum2024differentially}
Deniz Daum and et~al.
\newblock On differentially private 3d medical image synthesis with controllable latent diffusion models.
\newblock {\em MICCAI Workshop on Deep Generative Models}, 2024.

\bibitem[\protect\citeauthoryear{Deng \bgroup \em et al.\egroup }{2024}]{deng2024prdp}
Fei Deng, Qifei Wang, and et~al.
\newblock Prdp: Proximal reward difference prediction for large-scale reward finetuning of diffusion models.
\newblock {\em CVPR}, 2024.

\bibitem[\protect\citeauthoryear{Dong \bgroup \em et al.\egroup }{2023}]{dong2023aligndiff}
Zibin Dong, Yifu Yuan, and et~al.
\newblock Aligndiff: Aligning diverse human preferences via behavior-customisable diffusion model.
\newblock {\em ArXiv}, 2023.

\bibitem[\protect\citeauthoryear{Du \bgroup \em et al.\egroup }{2023}]{du2023beyond}
Hongyang Du, Ruichen Zhang, and et~al.
\newblock Beyond deep reinforcement learning: A tutorial on generative diffusion models in network optimization.
\newblock {\em ArXiv}, 2023.

\bibitem[\protect\citeauthoryear{Fan \bgroup \em et al.\egroup }{2024}]{fan2024reinforcement}
Ying Fan, Olivia Watkins, and et~al.
\newblock Reinforcement learning for fine-tuning text-to-image diffusion models.
\newblock {\em Neurips}, 2024.

\bibitem[\protect\citeauthoryear{Feng \bgroup \em et al.\egroup }{2025}]{feng2025survey}
Tuo Feng, Wenguan Wang, and et~al.
\newblock A survey of world models for autonomous driving.
\newblock {\em ArXiv}, 2025.

\bibitem[\protect\citeauthoryear{Gao and et al.}{2023}]{gao2023can}
Jialu Gao and et~al.
\newblock Can pre-trained text-to-image models generate visual goals for reinforcement learning?
\newblock {\em Neurips}, 2023.

\bibitem[\protect\citeauthoryear{Guan and et al.}{2024}]{guan2024world}
Yanchen Guan and et~al.
\newblock World models for autonomous driving: An initial survey.
\newblock {\em T-IV}, 2024.

\bibitem[\protect\citeauthoryear{Guo \bgroup \em et al.\egroup }{2024}]{guo2024versat2i}
Jianshu Guo, Wenhao Chai, and et~al.
\newblock Versat2i: Improving text-to-image models with versatile reward.
\newblock {\em ArXiv}, 2024.

\bibitem[\protect\citeauthoryear{Han \bgroup \em et al.\egroup }{2024}]{han2024advancing}
Woojung Han, Chanyoung Kim, and et~al.
\newblock Advancing text-driven chest x-ray generation with policy-based reinforcement learning.
\newblock {\em MICCAI}, 2024.

\bibitem[\protect\citeauthoryear{He \bgroup \em et al.\egroup }{2023}]{he2023learning}
Huiguo He, Tianfu Wang, and et~al.
\newblock Learning profitable nft image diffusions via multiple visual-policy guided reinforcement learning.
\newblock {\em ACMMM}, 2023.

\bibitem[\protect\citeauthoryear{Ho \bgroup \em et al.\egroup }{2020}]{ho2020denoising}
Jonathan Ho, Ajay Jain, and et~al.
\newblock Denoising diffusion probabilistic models.
\newblock {\em Neurips}, 2020.

\bibitem[\protect\citeauthoryear{Huang \bgroup \em et al.\egroup }{2024a}]{DBLP:journals/corr/abs-2402-17525}
Yi~Huang, Jiancheng Huang, and et~al.
\newblock Diffusion model-based image editing: {A} survey.
\newblock {\em Arxiv}, 2024.

\bibitem[\protect\citeauthoryear{Huang \bgroup \em et al.\egroup }{2024b}]{huang2024gen}
Zhiyu Huang, Xinshuo Weng, and et~al.
\newblock Gen-drive: Enhancing diffusion generative driving policies with reward modeling and reinforcement learning fine-tuning.
\newblock {\em ArXiv}, 2024.

\bibitem[\protect\citeauthoryear{Janner \bgroup \em et al.\egroup }{2022}]{janner2022planningdiffusionflexiblebehavior}
Michael Janner, Yilun Du, and et~al.
\newblock Planning with diffusion for flexible behavior synthesis.
\newblock {\em ArXiv}, 2022.

\bibitem[\protect\citeauthoryear{Lahlou \bgroup \em et al.\egroup }{2023}]{lahlou2023theory}
Salem Lahlou, Tristan Deleu, and et~al.
\newblock A theory of continuous generative flow networks.
\newblock {\em ICML}, 2023.

\bibitem[\protect\citeauthoryear{Lee \bgroup \em et al.\egroup }{2023}]{lee2023aligning}
Kimin Lee, Hao Liu, and et~al.
\newblock Aligning text-to-image models using human feedback.
\newblock {\em ArXiv}, 2023.

\bibitem[\protect\citeauthoryear{Lee \bgroup \em et al.\egroup }{2025}]{lee2025parrot}
Seung~Hyun Lee, Yinxiao Li, and et~al.
\newblock Parrot: Pareto-optimal multi-reward reinforcement learning framework for text-to-image generation.
\newblock {\em ECCV}, 2025.

\bibitem[\protect\citeauthoryear{Li \bgroup \em et al.\egroup }{2023}]{li2023drivingdiffusion}
Xiaofan Li, Yifu Zhang, and et~al.
\newblock Drivingdiffusion: Layout-guided multi-view driving scene video generation with latent diffusion model.
\newblock {\em ArXiv}, 2023.

\bibitem[\protect\citeauthoryear{Li \bgroup \em et al.\egroup }{2024a}]{li2024aligningdiffusionmodelsoptimizing}
Shufan Li, Konstantinos Kallidromitis, and et~al.
\newblock Aligning diffusion models by optimizing human utility.
\newblock {\em ArXiv}, 2024.

\bibitem[\protect\citeauthoryear{Li \bgroup \em et al.\egroup }{2024b}]{li2024instructrl4pix}
Tiancheng Li, Jinxiu Liu, and et~al.
\newblock Instructrl4pix: Training diffusion for image editing by reinforcement learning.
\newblock {\em ArXiv}, 2024.

\bibitem[\protect\citeauthoryear{Li \bgroup \em et al.\egroup }{2024c}]{li2024textcraftor}
Yanyu Li, Xian Liu, and et~al.
\newblock Textcraftor: Your text encoder can be image quality controller.
\newblock {\em CVPR}, 2024.

\bibitem[\protect\citeauthoryear{Liang \bgroup \em et al.\egroup }{2024}]{liang2024rich}
Youwei Liang, Junfeng He, and et~al.
\newblock Rich human feedback for text-to-image generation.
\newblock {\em CVPR}, 2024.

\bibitem[\protect\citeauthoryear{Liao \bgroup \em et al.\egroup }{2024}]{liao2024diffusiondrive}
Bencheng Liao, Shaoyu Chen, and et~al.
\newblock Diffusiondrive: Truncated diffusion model for end-to-end autonomous driving.
\newblock {\em ArXiv}, 2024.

\bibitem[\protect\citeauthoryear{Lin \bgroup \em et al.\egroup }{2023}]{lin2023magic3d}
Chen-Hsuan Lin, Jun Gao, and et~al.
\newblock Magic3d: High-resolution text-to-3d content creation.
\newblock {\em CVPR}, 2023.

\bibitem[\protect\citeauthoryear{Liu \bgroup \em et al.\egroup }{2024a}]{liu2024ddm}
Jiaqi Liu, Peng Hang, and et~al.
\newblock Ddm-lag: A diffusion-based decision-making model for autonomous vehicles with lagrangian safety enhancement.
\newblock {\em ArXiv}, 2024.

\bibitem[\protect\citeauthoryear{Liu \bgroup \em et al.\egroup }{2024b}]{liu2024rdt}
Songming Liu, Lingxuan Wu, and et~al.
\newblock Rdt-1b: a diffusion foundation model for bimanual manipulation.
\newblock {\em ArXiv}, 2024.

\bibitem[\protect\citeauthoryear{Ma{\~n}as and et al.}{2024}]{manas2024improving}
Oscar Ma{\~n}as and et~al.
\newblock Improving text-to-image consistency via automatic prompt optimization.
\newblock {\em ArXiv}, 2024.

\bibitem[\protect\citeauthoryear{Menapace \bgroup \em et al.\egroup }{2024}]{menapace2024promptable}
Willi Menapace, Aliaksandr Siarohin, and et~al.
\newblock Promptable game models: Text-guided game simulation via masked diffusion models.
\newblock {\em TOG}, 2024.

\bibitem[\protect\citeauthoryear{Miao and et al.}{2024}]{miao2024training}
Zichen Miao and et~al.
\newblock Training diffusion models towards diverse image generation with rl.
\newblock {\em CVPR}, 2024.

\bibitem[\protect\citeauthoryear{Mo \bgroup \em et al.\egroup }{2024}]{mo2024dynamic}
Wenyi Mo, Tianyu Zhang, and et~al.
\newblock Dynamic prompt optimizing for text-to-image generation.
\newblock {\em CVPR}, 2024.

\bibitem[\protect\citeauthoryear{Oertell \bgroup \em et al.\egroup }{2024}]{oertell2024rl}
Owen Oertell, Jonathan~Daniel Chang, and et~al.
\newblock Rl for consistency models: Reward guided text-to-image generation with fast inference.
\newblock {\em RLC}, 2024.

\bibitem[\protect\citeauthoryear{Ogezi and Shi}{2024}]{ogezi2024optimizing}
Michael Ogezi and Ning Shi.
\newblock Optimizing negative prompts for enhanced aesthetics and fidelity in text-to-image generation.
\newblock {\em ArXiv}, 2024.

\bibitem[\protect\citeauthoryear{Prabhudesai and et al.}{2024}]{prabhudesai2024aligningtexttoimagediffusionmodels}
Mihir Prabhudesai and et~al.
\newblock Aligning text-to-image diffusion models with reward backpropagation, 2024.

\bibitem[\protect\citeauthoryear{Qin and Weaver}{2024}]{10765093}
Xue Qin and Garrett Weaver.
\newblock Utilizing generative ai for vr exploration testing: A case study.
\newblock {\em ASEW}, 2024.

\bibitem[\protect\citeauthoryear{Rafailov \bgroup \em et al.\egroup }{2024}]{rafailov2024direct}
Rafael Rafailov, Archit Sharma, and et~al.
\newblock Direct preference optimization: Your language model is secretly a reward model.
\newblock {\em Neurips}, 2024.

\bibitem[\protect\citeauthoryear{Rahman \bgroup \em et al.\egroup }{2023}]{rahman2023ambiguous}
Aimon Rahman, Jeya Maria~Jose Valanarasu, and et~al.
\newblock Ambiguous medical image segmentation using diffusion models.
\newblock {\em CVPR}, 2023.

\bibitem[\protect\citeauthoryear{Ramesh \bgroup \em et al.\egroup }{2022}]{ramesh2022hierarchical}
Aditya Ramesh, Prafulla Dhariwal, Alex Nichol, Casey Chu, and Mark Chen.
\newblock Hierarchical text-conditional image generation with clip latents.
\newblock {\em ArXiv}, 2022.

\bibitem[\protect\citeauthoryear{Ren \bgroup \em et al.\egroup }{2025}]{ren2025refiningalignmentframeworkdiffusion}
Jie Ren, Yuhang Zhang, and et~al.
\newblock Refining alignment framework for diffusion models with intermediate-step preference ranking.
\newblock {\em ArXiv}, 2025.

\bibitem[\protect\citeauthoryear{Saharia \bgroup \em et al.\egroup }{2022}]{saharia2022photorealistic}
Chitwan Saharia, William Chan, and et~al.
\newblock Photorealistic text-to-image diffusion models with deep language understanding.
\newblock {\em Neurips}, 2022.

\bibitem[\protect\citeauthoryear{Schulman and et al.}{2015}]{schulman2015high}
John Schulman and et~al.
\newblock High-dimensional continuous control using generalized advantage estimation.
\newblock {\em ArXiv}, 2015.

\bibitem[\protect\citeauthoryear{Schulman \bgroup \em et al.\egroup }{2017}]{schulman2017proximal}
John Schulman, Filip Wolski, and et~al.
\newblock Proximal policy optimization algorithms.
\newblock {\em ArXiv}, 2017.

\bibitem[\protect\citeauthoryear{Song \bgroup \em et al.\egroup }{2020}]{song2020denoising}
Jiaming Song, Chenlin Meng, and et~al.
\newblock Denoising diffusion implicit models.
\newblock {\em ArXiv}, 2020.

\bibitem[\protect\citeauthoryear{Uehara \bgroup \em et al.\egroup }{2024a}]{uehara2024finetuningcontinuoustimediffusionmodels}
Masatoshi Uehara, Yulai Zhao, and et~al.
\newblock Fine-tuning of continuous-time diffusion models as entropy-regularized control.
\newblock {\em ArXiv}, 2024.

\bibitem[\protect\citeauthoryear{Uehara \bgroup \em et al.\egroup }{2024b}]{DBLP:journals/corr/abs-2407-13734}
Masatoshi Uehara, Yulai Zhao, and et~al.
\newblock Understanding reinforcement learning-based fine-tuning of diffusion models: {A} tutorial and review.
\newblock {\em Arxiv}, 2024.

\bibitem[\protect\citeauthoryear{Wallace and et al.}{2024}]{wallace2024diffusion}
Bram Wallace and et~al.
\newblock Diffusion model alignment using direct preference optimization.
\newblock {\em CVPR}, 2024.

\bibitem[\protect\citeauthoryear{Wang \bgroup \em et al.\egroup }{2023}]{wang2023diffusebot}
Tsun-Hsuan~Johnson Wang, Juntian Zheng, and et~al.
\newblock Diffusebot: Breeding soft robots with physics-augmented generative diffusion models.
\newblock {\em Neurips}, 2023.

\bibitem[\protect\citeauthoryear{Wang \bgroup \em et al.\egroup }{2024a}]{wang2024diffchat}
Jiapeng Wang, Chengyu Wang, and et~al.
\newblock Diffchat: Learning to chat with text-to-image synthesis models for interactive image creation.
\newblock {\em ArXiv}, 2024.

\bibitem[\protect\citeauthoryear{Wang \bgroup \em et al.\egroup }{2024b}]{wang2024sparse}
Yixiao Wang, Yifei Zhang, and et~al.
\newblock Sparse diffusion policy: A sparse, reusable, and flexible policy for robot learning.
\newblock {\em ArXiv}, 2024.

\bibitem[\protect\citeauthoryear{Wen and et al.}{2024}]{wen2024panacea}
Yuqing Wen and et~al.
\newblock Panacea: Panoramic and controllable video generation for autonomous driving.
\newblock {\em CVPR}, 2024.

\bibitem[\protect\citeauthoryear{Wen \bgroup \em et al.\egroup }{2023}]{wen2023improving}
Song Wen, Guian Fang, and et~al.
\newblock Improving compositional text-to-image generation with large vision-language models.
\newblock {\em ArXiv}, 2023.

\bibitem[\protect\citeauthoryear{Winata \bgroup \em et al.\egroup }{2024}]{winata2024preference}
Genta~Indra Winata, Hanyang Zhao, and et~al.
\newblock Preference tuning with human feedback on language, speech, and vision tasks: A survey.
\newblock {\em ArXiv}, 2024.

\bibitem[\protect\citeauthoryear{Wu and et al.}{2022}]{wu2022medsegdiff}
Junde Wu and et~al.
\newblock Medsegdiff: Medical image segmentation with diffusion probabilistic model.
\newblock {\em ArXiv}, 2022.

\bibitem[\protect\citeauthoryear{Wu and et al.}{2024}]{wu2024medsegdiff}
Junde Wu and et~al.
\newblock Medsegdiff-v2: Diffusion-based medical image segmentation with transformer.
\newblock {\em AAAI}, 2024.

\bibitem[\protect\citeauthoryear{Wu \bgroup \em et al.\egroup }{2023}]{wu2023human}
Xiaoshi Wu, Keqiang Sun, and et~al.
\newblock Human preference score: Better aligning text-to-image models with human preference.
\newblock {\em ICCV}, 2023.

\bibitem[\protect\citeauthoryear{Wu \bgroup \em et al.\egroup }{2024}]{wu2024protein}
Kevin~E Wu, Kevin~K Yang, and et~al.
\newblock Protein structure generation via folding diffusion.
\newblock {\em Nature communications}, 2024.

\bibitem[\protect\citeauthoryear{Wu \bgroup \em et al.\egroup }{2025}]{wu2025deep}
Xiaoshi Wu, Yiming Hao, and et~al.
\newblock Deep reward supervisions for tuning text-to-image diffusion models.
\newblock {\em ECCV}, 2025.

\bibitem[\protect\citeauthoryear{Xie and Li}{2022}]{xie2022measurement}
Yutong Xie and Quanzheng Li.
\newblock Measurement-conditioned denoising diffusion probabilistic model for under-sampled medical image reconstruction.
\newblock {\em MICCAI}, 2022.

\bibitem[\protect\citeauthoryear{Xu \bgroup \em et al.\egroup }{2024}]{xu2024imagereward}
Jiazheng Xu, Xiao Liu, and et~al.
\newblock Imagereward: Learning and evaluating human preferences for text-to-image generation.
\newblock {\em Neurips}, 2024.

\bibitem[\protect\citeauthoryear{Yang and et al.}{2024}]{yang2024using}
Kai Yang and et~al.
\newblock Using human feedback to fine-tune diffusion models without any reward model.
\newblock {\em CVPR}, 2024.

\bibitem[\protect\citeauthoryear{Yang \bgroup \em et al.\egroup }{2023}]{yang2023learning}
Mengjiao Yang, Yilun Du, and et~al.
\newblock Learning interactive real-world simulators.
\newblock {\em ArXiv}, 2023.

\bibitem[\protect\citeauthoryear{Yang \bgroup \em et al.\egroup }{2024a}]{yang2024mastering}
Ling Yang, Zhaochen Yu, and et~al.
\newblock Mastering text-to-image diffusion: Recaptioning, planning, and generating with multimodal llms.
\newblock {\em ICML}, 2024.

\bibitem[\protect\citeauthoryear{Yang \bgroup \em et al.\egroup }{2024b}]{yang2024dense}
Shentao Yang, Tianqi Chen, and Mingyuan Zhou.
\newblock A dense reward view on aligning text-to-image diffusion with preference.
\newblock {\em ArXiv}, 2024.

\bibitem[\protect\citeauthoryear{Yang \bgroup \em et al.\egroup }{2025}]{yang2025ipoiterativepreferenceoptimization}
Xiaomeng Yang, Zhiyu Tan, and et~al.
\newblock Ipo: Iterative preference optimization for text-to-video generation.
\newblock {\em ArXiv}, 2025.

\bibitem[\protect\citeauthoryear{Yu \bgroup \em et al.\egroup }{2024}]{yu2024regularized}
Xudong Yu, Chenjia Bai, and et~al.
\newblock Regularized conditional diffusion model for multi-task preference alignment.
\newblock {\em ArXiv}, 2024.

\bibitem[\protect\citeauthoryear{Yuan \bgroup \em et al.\egroup }{2024a}]{yuan2024self}
Huizhuo Yuan, Zixiang Chen, and et~al.
\newblock Self-play fine-tuning of diffusion models for text-to-image generation.
\newblock {\em ArXiv}, 2024.

\bibitem[\protect\citeauthoryear{Yuan \bgroup \em et al.\egroup }{2024b}]{yuan2024preferencealigneddiffusionplanner}
Xinyi Yuan, Zhiwei Shang, and et~al.
\newblock Preference aligned diffusion planner for quadrupedal locomotion control.
\newblock {\em ArXiv}, 2024.

\bibitem[\protect\citeauthoryear{Zhang \bgroup \em et al.\egroup }{2022}]{zhang2022unifying}
Dinghuai Zhang, Ricky~TQ Chen, and et~al.
\newblock Unifying generative models with gflownets and beyond.
\newblock {\em ArXiv}, 2022.

\bibitem[\protect\citeauthoryear{Zhang \bgroup \em et al.\egroup }{2023}]{DBLP:journals/corr/abs-2303-07909}
Chenshuang Zhang, Chaoning Zhang, and et~al.
\newblock Text-to-image diffusion models in generative {AI:} {A} survey.
\newblock {\em Arxiv}, 2023.

\bibitem[\protect\citeauthoryear{Zhang \bgroup \em et al.\egroup }{2024a}]{zhang2024improving}
Dinghuai Zhang, Yizhe Zhang, and et~al.
\newblock Improving gflownets for text-to-image diffusion alignment.
\newblock {\em ArXiv}, 2024.

\bibitem[\protect\citeauthoryear{Zhang \bgroup \em et al.\egroup }{2024b}]{zhang2024onlinevpoalignvideodiffusion}
Jiacheng Zhang, Jie Wu, and et~al.
\newblock Onlinevpo: Align video diffusion model with online video-centric preference optimization.
\newblock {\em ArXiv}, 2024.

\bibitem[\protect\citeauthoryear{Zhang \bgroup \em et al.\egroup }{2024c}]{zhang2024hive}
Shu Zhang, Xinyi Yang, and et~al.
\newblock Hive: Harnessing human feedback for instructional visual editing.
\newblock {\em CVPR}, 2024.

\bibitem[\protect\citeauthoryear{Zhang \bgroup \em et al.\egroup }{2025}]{zhang2025large}
Yinan Zhang, Eric Tzeng, and et~al.
\newblock Large-scale reinforcement learning for diffusion models.
\newblock {\em ECCV}, 2025.

\bibitem[\protect\citeauthoryear{Zhou \bgroup \em et al.\egroup }{2024}]{zhou2024surveygenerativeaillm}
Pengyuan Zhou, Lin Wang, and et~al.
\newblock A survey on generative ai and llm for video generation, understanding, and streaming.
\newblock {\em arXiv}, 2024.

\end{thebibliography}
